\documentclass[sigconf]{acmart}
\newcommand{\ignore}[1]{}
\usepackage{xspace}

\newcommand{\be}{\mathbf{e}}
\newcommand{\bm}{\mathbf{m}}
\newcommand{\bv}{\mathbf{v}}
\newcommand{\bu}{\mathbf{u}}
\newcommand{\bk}{\mathbf{k}}
\newcommand{\bg}{\mathbf{g}}
\newcommand{\bd}{\mathbf{d}}
\newcommand{\br}{\mathbf{r}}
\newcommand{\bb}{\mathbf{b}}
\newcommand{\bs}{\mathbf{s}}
\newcommand{\bh}{\mathbf{h}}
\newcommand{\bA}{\mathbf{A}}
\newcommand{\bH}{\mathbf{H}}
\newcommand{\bB}{\mathbf{B}}
\newcommand{\bE}{\mathbf{E}}
\newcommand{\bF}{\mathbf{F}}
\newcommand{\bW}{\mathbf{W}}
\newcommand{\bC}{\mathbf{C}}
\newcommand{\bM}{\mathbf{M}}
\newcommand{\bD}{\mathbf{D}}
\newcommand{\bP}{\mathbf{P}}
\newcommand{\bX}{\mathbf{X}}
\newcommand{\bQ}{\mathbf{Q}}
\newcommand{\bS}{\mathbf{S}}
\newcommand{\bK}{\mathbf{K}}
\newcommand{\bV}{\mathbf{V}}
\newcommand{\bR}{\mathbf{R}}
\let\Bbbk\relax
\usepackage{amsfonts,amssymb}
\usepackage{booktabs} 
\usepackage{caption}
\usepackage{subcaption}
\usepackage{bm}
\usepackage{multirow}
\usepackage{footmisc}
\usepackage{amssymb}
\usepackage{array}
\usepackage{graphicx}
\usepackage{clrscode}
\usepackage{balance}
\usepackage{float}
\usepackage{color}
\usepackage{xcolor}
\usepackage{amsopn}
\usepackage{mathrsfs}
\usepackage{mathtools}
\usepackage{amsmath}
\usepackage{arydshln}
\usepackage{hyperref}
\usepackage{blkarray}
\usepackage{enumerate}
\usepackage{courier}
\usepackage{mathrsfs}
\usepackage{rotating}
\usepackage{ragged2e}
\usepackage{multicol}
\usepackage{latexsym}
\usepackage{mflogo}
\usepackage{amsthm}
\usepackage{soul}
\usepackage{hhline}
\usepackage[ruled,linesnumbered]{algorithm2e}
\usepackage{makecell}
\usepackage{CJKutf8}
\newcommand{\paratitle}[1]{\vspace{1.5ex}\noindent\textbf{#1}}
\newcommand{\ie}{\emph{i.e.,}\xspace}
\newcommand{\aka}{\emph{a.k.a.,}\xspace}
\newcommand{\eg}{\emph{e.g.,}\xspace}
\newcommand{\etal}{\emph{et al.}\xspace}

\usepackage{epigraph}
\AtBeginDocument{%
	\providecommand\BibTeX{{%
			\normalfont B\kern-0.5em{\scshape i\kern-0.25em b}\kern-0.8em\TeX}}}

\copyrightyear{2022}
\acmYear{2022}
\setcopyright{acmcopyright}\acmConference[KDD '22]{Proceedings of the 28th ACM SIGKDD Conference on Knowledge Discovery and Data Mining}{August 14--18, 2022}{Washington, DC, USA}
\acmBooktitle{Proceedings of the 28th ACM SIGKDD Conference on Knowledge Discovery and Data Mining (KDD '22), August 14--18, 2022, Washington, DC, USA}
\acmPrice{15.00}
\acmDOI{10.1145/3534678.3539131}
\acmISBN{978-1-4503-9385-0/22/08}

\begin{document}
	
	\title{JiuZhang: A Chinese Pre-trained Language Model for\\ Mathematical Problem Understanding}
	
	\author{Wayne Xin Zhao}
	\authornotemark[2]
	\affiliation{%
		\institution{Gaoling School of Artificial Intelligence, Renmin University of China}
		\city{Beijing}
		\country{China}
	}\email{batmanfly@gmail.com}
	
	\author{Kun Zhou}
	\authornote{These four authors contributed equally to this research.} 
	\affiliation{%
		\institution{School of Information, Renmin University of China}
		\city{Beijing}
		\country{China}
	}\email{francis_kun_zhou@163.com}
	
	\author{Zheng Gong}
	\authornotemark[1]
	\affiliation{%
		\institution{Gaoling School of Artificial Intelligence, Renmin University of China}
		\city{Beijing}
		\country{China}
	}\email{gongzheng0109@ruc.edu.cn}
	
	\author{Beichen Zhang}
	\authornotemark[1]
	\affiliation{%
		\institution{Gaoling School of Artificial Intelligence, Renmin University of China}
		\city{Beijing}
		\country{China}
	}\email{zhangbeichen724@gmail.com}
	
	\author{Yuanhang Zhou}
	\authornotemark[1]
	\affiliation{%
		\institution{School of Information, Renmin University of China}
		\city{Beijing}
		\country{China}
	}\email{sdzyh002@gmail.com}
	
	\author{Jing Sha}
	\affiliation{%
		\institution{iFLYTEK Research}
		\institution{State Key Laboratory of Cognitive Intelligence}
		\city{Hefei}
		\country{China}}
	\email{jingsha@iflytek.com}
	
	\author{Zhigang Chen}
	\affiliation{%
		\institution{iFLYTEK Research}
		\institution{State Key Laboratory of Cognitive Intelligence}
		\city{Hefei}
		\country{China}}
	\email{zgchen@iflytek.com}
	
	\author{Shijin Wang}
	\affiliation{%
		\institution{AI Research~(Central China), iFLYTEK}
		\institution{State Key Laboratory of Cognitive Intelligence}
		\city{Wuhan}
		\country{China}}
	\email{sjwang3@iflytek.com}
	
	\author{Cong Liu}
	\affiliation{%
		\institution{iFLYTEK Research}
		\city{Hefei}
		\country{China}}
	\email{congliu2@iflytek.com}
	
	\author{Ji-Rong Wen}
	\authornote{Also with Beijing Key Laboratory of Big Data Management and Analysis Methods.}
	\affiliation{%
		\institution{Gaoling School of Artificial Intelligence, Renmin University of China}
		\city{Beijing}
		\country{China}
	}\email{jrwen@ruc.edu.cn}
	
	\ignore{
		\author{Wayne Xin Zhao$^{1,3}$, Kun Zhou$^{2\dagger}$, Zheng Gong$^{1\dagger}$, Beichen Zhang$^{1\dagger}$, Yuanhang Zhou$^{2\dagger}$,\\  Jing Sha$^{4}$, Zhigang Chen$^{4}$, Shijin Wang$^{4}$, Cong Liu$^{4}$, Ji-Rong Wen$^{1,3}$}\thanks{$^\dagger$Equal contribution} 
		\affiliation{%
			\institution{$^1$Gaoling School of Artificial Intelligence, Renmin University of China, Beijing, China}
			\institution{$^2$School of Information, Renmin University of China, Beijing, China}
			\institution{$^3$Beijing Key Laboratory of Big Data Management and Analysis Methods, Beijing, China}
			\institution{$^4$iFLYTEK Research, Hefei, China}
			\country{}
	}}
	\renewcommand{\shortauthors}{Zhao, et al.}
	\begin{abstract}
		This paper aims to advance the mathematical intelligence of machines by presenting the first Chinese mathematical pre-trained language model~(PLM) for effectively understanding and representing mathematical problems. Unlike other standard NLP tasks, mathematical texts are difficult to understand, since they involve mathematical terminology, symbols and formulas in the problem statement. Typically, it requires complex mathematical logic and background knowledge for solving mathematical problems.
		
		Considering the complex nature of mathematical texts, we design a novel curriculum pre-training approach for improving the learning of mathematical PLMs, consisting of both basic and advanced courses. 
		Specially, we first perform token-level pre-training based on a position-biased masking strategy, and then design logic-based pre-training tasks that aim to recover the shuffled sentences and formulas, respectively. Finally, we introduce a more difficult pre-training task that enforces the PLM to detect and correct the errors in its generated solutions.  We conduct extensive experiments on offline evaluation  (including nine math-related tasks) and online $A/B$ test.
		Experimental results demonstrate the effectiveness of our approach compared with a number of competitive baselines.
		Our code is available at: \textcolor{blue}{\url{https://github.com/RUCAIBox/JiuZhang}}.
	\end{abstract}

	
	
	
	\maketitle

	\section{Introduction}
	
	\epigraph{Pure mathematics is, in its way, the poetry of logical ideas.}{\textit{Albert Einstein}}

	Mathematical ability is a human construct, referring to  the ability to obtain, process, and retain mathematical information from the cognitive perspective~\cite{Karsenty2014}. 
	It has been widely recognized that it is difficult for machines to grasp such an intelligent ability via computational models, which requires a wide spectrum of mathematical knowledge, logic and skills.
	Existing methods in the literature usually adopt a natural language processing~(NLP) approach, focusing on analyzing and understanding the semantics in the mathematical texts (\ie text that presents the problem statement or answer key).
	Early studies design explicit extraction methods based on statistical features~\cite{fletcher1985understanding}, semantic parser~\cite{shi2015automatically} and other heuristic methods~\cite{kushman2014learning}. In contrast, recent progress focuses on learning latent semantic representations for mathematical texts with deep neural networks~\cite{wang2017deep,chiang2019semantically,li2019modeling}. However, due to complex mathematical logic and knowledge, it is still challenging to accurately understand mathematical problems, which is the fundamental step to develop mathematical intelligence for machines. 
	
	Unlike other standard NLP tasks (\eg part-of-speech tagging and named entity recognition), it is more difficult to understand mathematical texts, since they mix mathematical terminology, symbols and formulas in the problem statement, requiring complex mathematical logic and background knowledge for deriving the solution.
	Recently,  pre-trained language models (PLMs)~\cite{devlin-etal-2019-bert,peng2021mathbert} have shed light on more effective approaches to the understanding and representation of mathematical texts. 
	After being pre-trained on large-scale general corpus, PLMs can encode rich semantic knowledge and linguistic characteristics by massive parameters of the Transformer architecture~\cite{rogers-etal-2020-primer}.
	Furthermore, they can deal with downstream  tasks via fine-tuning or continual pre-training~\cite{gururangan-etal-2020-dont,gong-etal-2022-continual}. When adapting to mathematical texts, existing methods~\cite{peng2021mathbert,shen2021mathbert}  either adopt the masked language model task (MLM) to improve the understanding of mathematical symbols and  terminology~\cite{shen2021mathbert}, or devise specific  pre-training tasks to enhance the semantic relatedness  between text description and formulas~\cite{peng2021mathbert,gong-etal-2022-continual}. 
	
	Although PLMs have achieved remarkable performance on preliminary mathematical tasks, they can't perform sufficiently well on higher-level tasks that require a more deep understanding of advanced mathematical knowledge and logic, \eg  high-school multi-choice questions and proof problems.
	A major reason is that the adopted pre-training tasks (\eg the MLM task) mainly capture \emph{textual semantics} via contextual  co-occurrences instead of  \emph{mathematical semantics} via complex mathematical knowledge or logic.
	As a result, these PLMs might produce a  linguistically reasonable but mathematically incorrect answer to a mathematical problem (\eg one plus one equals to \underline{\emph{three}}), since it is not fully aware of the underlying mathematical semantics. 
	To accurately understand the mathematical semantics, it is essential to develop more effective pre-training tasks or strategies for enhancing the representation capacity of PLMs on mathematical texts.
	As another major limit, existing mathematical PLMs are mostly pre-trained on English corpus, which is not directly applicable to non-English domains.

	To address these issues, in this paper, we propose the first  Chinese PLM for mathematical problem understanding, named as \textbf{JiuZhang}\footnote{Named after one classical Chinese mathematical book.}.  JiuZhang is developed based on the Transformer~\cite{vaswani2017attention} architecture,  consisting  of a shared Transformer encoder, a decoder for the understanding tasks ($U$-decoder) and a decoder for the generation tasks ($G$-decoder), which endows the model with the flexibility to deal with different downstream tasks.
	To pre-train JiuZhang, we collect a large-scale Chinese corpus consisting of 1,276,952 high-school mathematical exercises or tests, covering a variety of problem types such as multi-choice and blank-filling problems.
	
	As the major technical contribution, we design a curriculum pre-training approach to improving the understanding of mathematical knowledge and logic,  from \emph{basic} to \emph{advanced} courses. As the basic course, we aim to enhance the understanding of math symbols and their semantic relatedness with the text, and design a position-biased masking strategy that assigns larger masking probabilities to the tokens nearer to the answer in the solution text. As the advanced courses, we focus on improving the capacity of mathematical logic reasoning and solution checking, which are essential to solve complex math tasks.  
	For mathematical logic reasoning, we introduce the reconstruction tasks that recover the shuffled sentences or formulas in the mathematical text. For solution checking, we enforce the two decoders to detect and correct the incorrect generation contents from each other.  
	Such a pre-training process resembles the student-learning progress~\cite{10.1145/1553374.1553380}, where she/he gradually grasps a knowledge point from symbols, logic to problem solving.

	To the best of our knowledge, it is the first Chinese PLM specially for mathematical problem understanding. We conduct experiments on nine tasks from high-school math education, including basic classification tasks, math text retrieval tasks, question answering tasks and solution generation tasks.
	Experimental results have shown that our approach outperforms eleven baseline models, including competitive Chinese PLMs.
	Besides, we also deploy our model in the Zhixue app and online $A/B$ tests further verify the effectiveness of our approach. 

	\ignore{In order to improve the understanding of Chinese math problems, our solutions are twofold. 
		First, based on the large-scale Chinese math problem corpus, we design a set of pre-training tasks to both improve the understanding of the math text and symbols and capture the essential mathematical logic.
		Concretely, we adopt the masked language model (MLM) and denoised auto-encoding (DAE) tasks to learn the context-token/symbol correlation, the logic-related tasks that recover the shuffled sentences or formulas to learn the coarse-grained casual logic, the self-correction tasks that detect and correct the generated wrong tokens to learn the fine-grained mathematical logic.
		Second, we organize the learning order of the above tasks and devise a curriculum pre-training approach to effectively improve the understanding of basic and advanced mathematical knowledge and logic.
		Inspired by the math education process of human, we design the curriculum that aims to learn from basic knowledge (MLM and DAE) to advanced logic (logic-related tasks) to self error correction (self-correction tasks).
		In this way, the model is able to better utilize the basic knowledge to improve the learning of advanced logic, and then boosts itself by learning from its errors.
		To this end, we propose \textbf{JiuZhang}~\footnote{Named from one of the earliest Chinese mathematics book.}, a Chinese PLM for mathematical problem understanding.
		JiuZhang is based on the architecture of CPT~\cite{shao2021cpt} that consists of a shared Transformer encoder, a decoder for accomplishing understanding tasks (U-Decoder) and a decoder for generation tasks (G-Decoder), which endow the generality of our model on a variety of downstream tasks.
		Given the unsupervised Chinese math problem corpus, we first adopt the MLM and DAE tasks to learn the parameters of the U-Decoder and G-Decoder, respectively, where we also adopt a linear probability mask strategy that enforces the model to focus more on the deduced analysis and answer part.
		Then, we further pre-train the model parameters with the logic-related tasks to further improve the understanding of mathematical logic, which learn to recover the shuffled sentences and formulas.
		Finally, we continually pre-train the model by self-correction learning, where we first mask part of tokens from the analysis of a math problem and then utilize the U-decoder/G-decoder to detect and correct the predicted masked tokens of the G-decoder/U-decoder.
		We conduct extensive experiments on nine tasks in the math domain, including basic classification tasks, math text retrieval tasks, question answering tasks and analysis generation tasks.
		Experimental results have demonstrated the effectiveness of our approach.
	}
	\ignore{
		Our contributions can be summarized as follows:
		
		(1) We propose JiuZhang, a Chinese pre-training model for mathematical problem understanding.
		
		(2) We design a curriculum pre-training strategy to gradually improve the understanding of mathematical knowledge and logic.
		
		(3) Experiments on nine tasks in the math domain demonstrate the effectiveness of our model.
	}
	
	\section{Related Work}
	
	\paratitle{Mathematical Problem Understanding.}
	It is a key capacity to understand mathematical problems when developing artificial intelligence algorithms for 
	math education applications,  \eg mathematical problem retrieval~\cite{zanibbi2012recognition}, solving~\cite{lan2021mwptoolkit} and classification~\cite{peng2021mathbert}.
	As a fundamental step, it is essential to learn effective representations for mathematical texts, which involve symbols, formulas and texts that describe the  problem.
	Early methods adopt rule-based methods~\cite{fletcher1985understanding} to extract features for understanding the text and formulas, \eg semantic parser~\cite{shi2015automatically}, operator tree~\cite{zanibbi2012recognition} and variable entity alignment~\cite{kushman2014learning}.
	With the development of deep learning, 
	a surge of works adopt deep neural networks to develop more effective approaches for mathematical problem understanding.
	For math word problems, RNN~\cite{chiang2019semantically} and Transformer~\cite{li2019modeling} have been utilized to encode the mathematical text and generate the math equation.
	For mathematical information retrieval, graph neural networks~\cite{song2021searching} have been adopted to learn meaningful  representations over the structured formulas.
	Recently, the success of      PLMs~\cite{peng2021mathbert,polu2020generative} pushes forward the understanding and modeling of mathematical texts, due to the excellent capacity of language modeling.
	For example, in order to enhance the understanding of complex math formulas and logic,  MathBERT~\cite{peng2021mathbert} and COMUS~\cite{gong-etal-2022-continual} pre-train BERT on a large-scale mathematical corpus with formula-related pre-training tasks.
	However, existing PLMs still rely on (or simply adapt) original pre-training tasks of BERT, which lack a full consideration of the characteristics of mathematical texts. 
	More recently, OpenAI proposes GPT-$f$ for automated theorem proving~\cite{polu2020generative}, and also uses the proofs generated by GPT-$f$ to iteratively improve its performance~\cite{polu2022formal}.
	Despite the remarkable performance, these methods require a million scale of parameters, which is not easy to be deployed or applied in real-world applications.
	
	\paratitle{Chinese Pre-trained Language Models.}
	Pre-trained Language Models~(PLMs) (\eg BERT~\cite{devlin-etal-2019-bert}, BART~\cite{lewis-etal-2020-bart} and T5~\cite{raffel2020exploring}) have largely advanced the progress of language intelligence.
	Following this direction, our work is based on Chinese PLMs. 
	The first line of works adapt BERT~\cite{devlin-etal-2019-bert} by reusing masked language model~(MLM) task to pre-train Transformer encoders~\cite{cui2021pre,sun2021chinesebert} on Chinese corpus. They consider modeling the linguistic characteristics or semantic knowledge of Chinese texts and devise special strategies to improve the task performance, \eg whole word masking~\cite{cui2021pre}, glyph and pinyin embedding~\cite{sun2021chinesebert} and entity enhanced embedding~\cite{jia-etal-2020-entity}. Focused on natural language understanding, these methods can't be directly applied to text generation tasks.
	Another line of works pre-train Transformer via the auto-regressive task~\cite{zhang2021cpm,zeng2021pangu} or the seq2seq task~\cite{zhang2021mengzi}. They predict or recover the corrupted tokens  from left to right.
	Furthermore,  several studies attempt to endow Chinese PLMs with both capacities of understanding and generation. For example,
	Mengzi~\cite{zhang2021mengzi} and ERNIE-3.0~\cite{DBLP:journals/corr/abs-2107-02137}  consist of shared encoding layers and multiple task-specific decoders.
	With a similar architecture, CPT~\cite{shao2021cpt} adopts a deeper encoder and two shallower decoders, to accelerate the inference of text generation.  Our work presents the first Chinese PLM for understanding mathematical texts, with a series of specially designed pre-training tasks. 
	

	\section{Approach}
	In this section, we present the proposed PLM \textbf{JiuZhang} for mathematical problem understanding.
	We first formally describe the mathematical text as the pre-training corpus, then introduce the model architecture (consisting of a shared encoder and two task-specific decoders) and the curriculum pre-training approach (gradually learning the mathematical knowledge and logic from the mathematical corpus).
	Finally, we present the learning and discussion.
	
	\begin{figure*}
		\centering
		\includegraphics[scale=0.43]{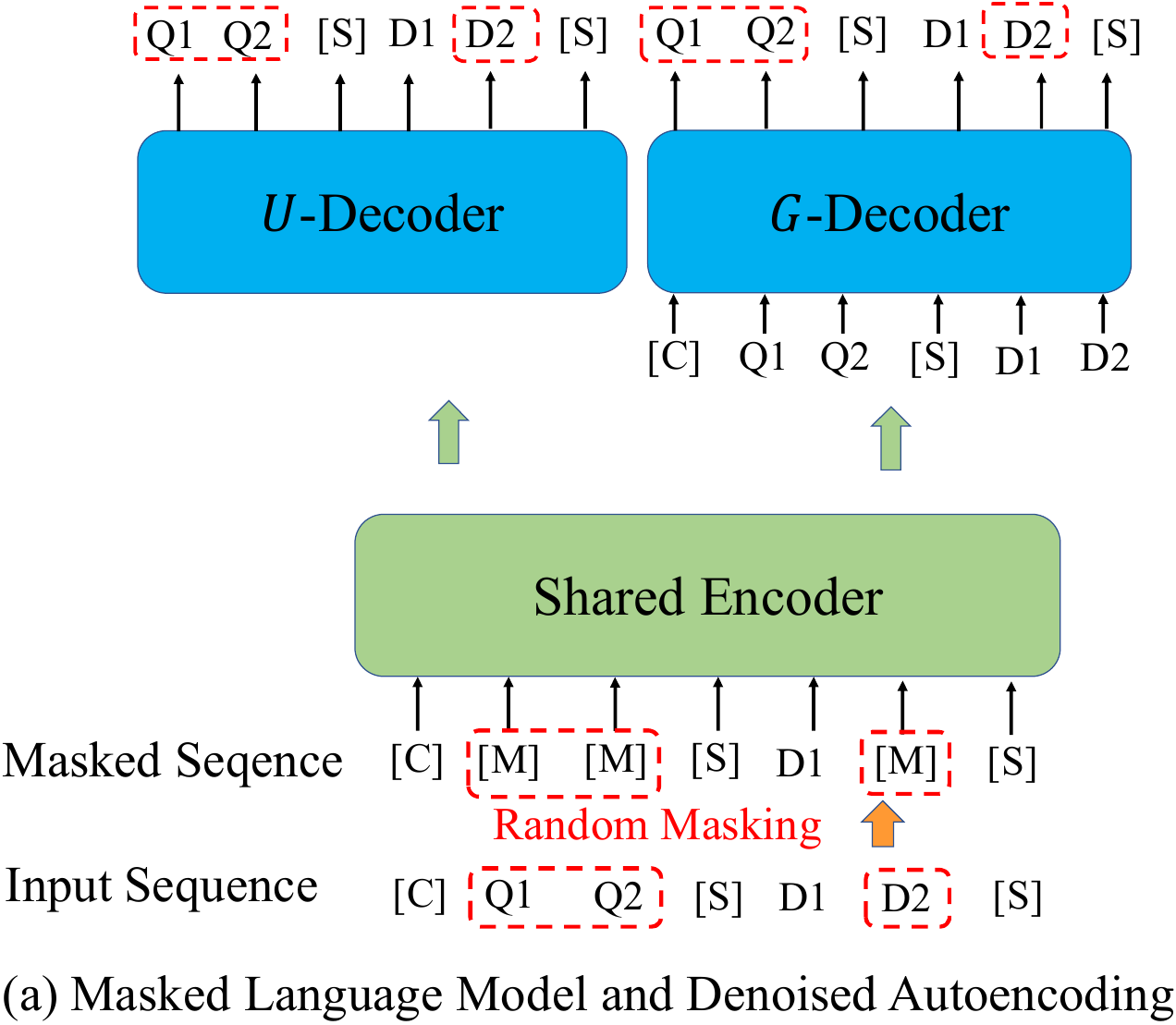}
		\hspace{0.1in}
		\includegraphics[scale=0.43]{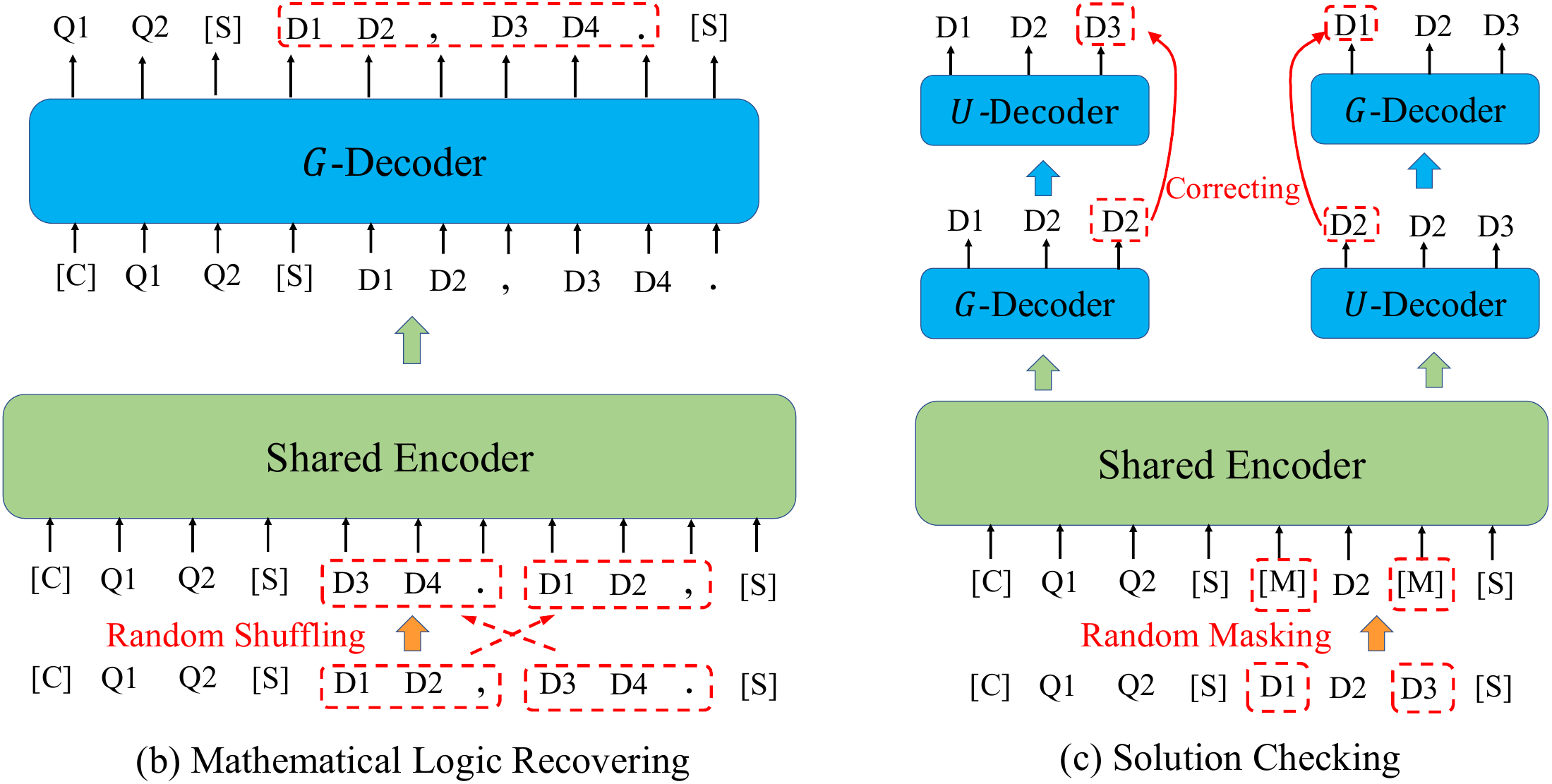}
		\caption{The overview of our curriculum pre-training approach, consisting of a basic course about masked token prediction, and two advanced courses about mathematical logic recovering and solution checking.}
		\label{approach}
	\end{figure*}

	\subsection{Mathematical Text}
	We first formally describe  \emph{mathematical text},  a general phrase referring  to the text relating to a mathematical problem. 
	In our corpus,  a mathematical problem is associated with two texts, namely problem statement and solution text (\aka answer key). 
	The problem statement introduces necessary background information and explains the problem that is to be solved, and  the answer key describes the key hints or complete solving  procedure for deriving the answer. 
	We concatenate both  problem statement and solution description as the \emph{mathematical text} of a mathematical problem. 
	
	Overall, a mathematical text is a  textual description that mixes text words with math symbols. Given a mathematical problem $q$, the corresponding mathematical text can be considered as a sequence of $n$ tokens, denoted as $q=\{t_1, t_2, \cdots , t_n\}$, where each token $t_i$ is either a text word or a math symbol (denoting a variable or an operator).  Furthermore, a consecutive segment of $l$ math symbols constitute a math formula, denoted as $f_i = \{s_1, \cdots, s_l\}$, where each math symbol $s_j$ is from $\{t_1, t_2, \cdots , t_n\}$. There are usually multiple formulas in a mathematical text, denoted as $\{f_1, f_2, \cdots , f_m\}$.
	
	Based on the above notations, this work focuses on pre-training a PLM on a large-scale corpus consisting of  
	mathematical texts. Then, the PLM can be 
	fine-tuned on various mathematical tasks (\eg knowledge point classification and similar question retrieval), and improve the corresponding task performance.

	\subsection{Model Architecture}
	
	For dealing with various mathematical tasks, we borrow the idea of setting task-specific decoders from CPT~\cite{shao2021cpt} and Mengzi~\cite{zhang2021mengzi}. 
	The base architecture of JiuZhang consists of a shared Transformer encoder, a decoder for understanding tasks ($U$-decoder) and a decoder for generation tasks ($G$-decoder).  The two decoders endow our model with more flexibility to solve different downstream tasks. 
	
	\subsubsection{Shared Transformer Encoder}
	Given a mathematical problem $q=\{t_1, \cdots, t_{n}\}$, we adopt a shared Transformer encoder to learn contextualized token representations, which can be further utilized by $U$-decoder or $G$-decoder for pre-training or fine-tuning. 
	Following the standard Transformer~\cite{vaswani2017attention}, our encoder is composed of an embedding layer and multiple Transformer layers.
	In the embedding layer, we maintain a token embedding matrix and a position embedding matrix, and then project the mathematical problem into dense representations as $\mathbf{E}_{T} \in \mathbb{R}^{n \times k}$ and $\mathbf{E}_{P} \in \mathbb{R}^{n \times k}$, where $k$ is the embedding dimensionality. Note that math symbols are considered as special word tokens for embedding.  
	Then, we sum the two embedding matrices as $\mathbf{E}=\mathbf{E}_{T}+\mathbf{E}_{P}$.
	Based on the embedding layer, we stack $L$ Transformer layers (consisting of a multi-head self-attention layer and a point-wise feed-forward network) to further encode these contextualized embeddings $\mathbf{E}$.
	After $L$ Transformer layers, the representations in the final layer $\{\mathbf{h}^{(L)}_{1}, \mathbf{h}^{(L)}_{2}, \cdots, \mathbf{h}^{(L)}_{n}\}$ are taken as the output of the shared Transformer encoder for subsequent decoders.

	\subsubsection{Understanding- and Generation-specific Decoders.} To solve different types of mathematical tasks, we design two task-specific decoders (\ie $U$-decoder and $G$-decoder).
	Following the unbalanced design~\cite{shao2021cpt}, the two decoders contain much fewer layers than the shared encoder (\ie 2 layers \emph{v.s.} 10 layers).
	In this way, the number of the involved parameters (including the encoder and one decoder) for a single task is still similar to the original scale of the PLM architecture~\cite{devlin-etal-2019-bert}, while the decoding efficiency for generation tasks can be largely improved.
	Besides, since we use more layers in the shared encoder, it can better capture the underlying semantics of mathematical texts.

	The $U$-decoder is a shallow bidirectional Transformer specially for the understanding tasks, which adopts the similar architecture of the shared encoder consisting of $L_U$ bidirectional Transformer layers. 
	Following BERT~\cite{devlin-etal-2019-bert}, we insert a special symbol ``\textsf{[CLS]}'' at the beginning of a sentence and utilize its representation at the last layer as the output of the $U$-decoder for understanding tasks. 
	
	The $G$-decoder is a shallow auto-regressive Transformer for generation tasks. It is composed by a stack of $L_G$ Transformer layers with the masked self-attention mechanism, which prevents attending to subsequent positions during the teacher-forcing training of generation tasks. Each Transformer layer of the $G$-decoder contains three sub-layers, namely a multi-head self-attention layer, a cross-attention layer over the output representations of the shared Transformer encoder, and a point-wise feed-forward network. It performs decoding to generate the sequence for generation tasks.
	
	\subsection{Curriculum Pre-training}
	Based on the above model architecture, we further propose a curriculum pre-training approach, which  helps the model gradually learn the complex mathematical knowledge and logic from large-scale mathematical corpus.  
	To achieve it, we first utilize the basic course that consists of the masked language model and denoised auto-encoding tasks to learn basic mathematical semantics. Then, we adopt the advanced course of mathematical logic recovering that reconstructs the shuffled sentences and formulas to improve the understanding of mathematical logic. Finally, we propose a more difficult solution checking task to enhance the capacity to detect and correct the incorrect generations by itself.
	
	\subsubsection{Basic Course:  Masked Token Prediction}
	\label{sec-MLM}
	In the basic course, we aim to enhance the understanding of math symbols and establish the semantic relatedness between text words and math symbols. 
	Following BERT~\cite{devlin-etal-2019-bert}, we predict the masked tokens based on the contextualized representations. 
	Since we have $U$-decoder and $G$-decoder, we adopt the masked language model (MLM) and denoised auto-encoding (DAE) as the pre-training tasks for the two decoders, respectively. 
	Given the input text, we randomly mask part of tokens from it, and then utilize the $U$-decoder to predict the masked tokens via the MLM task and utilize the $G$-decoder to reconstruct the original sentence via the DAE task. In order to adapt to mathematical texts, we further propose a new position-biased masking strategy to focus on more tokens at larger positions. 
	
	\paratitle{Masked Token Prediction}.  
	Given a mathematical problem $q$, we concatenate its question statement and solution text as a sequence, denoted as $x$.
	Then, we randomly select 15\% tokens of the input sequence for masking (including both words and symbols), in which 80\% ones are replaced by the token ``\textsf{[MASK]}'', 10\% ones are replaced by a random token and the rest 10\% ones remain unchanged.
	Formally, let $\widetilde{x}$ denote the masked sequence,  $V_{mask}$ denote the selected tokens, and the MLM and DAE losses are defined as:
	\begin{align}
		L_{MLM} &=\sum_{t_i \in V_{mask}}-\log{p(t_i | \widetilde{x};\Theta_{E}, \Theta_{U})},
		\label{eq-mlm}\\
		L_{DAE} &=\sum_{i}-\log{p(t_i|t_{<i}, \widetilde{x};\Theta_{E}, \Theta_{G})},
		\label{eq-dae}
	\end{align}
	where $p(t_i | \widetilde{x};\Theta_{E}, \Theta_{U})$ and $p(t_i|t_{<i};\widetilde{x};\Theta_{E}, \Theta_{G})$ denote the prediction probabilities of the token $t_i$ at the $i$-th position according to $U$-decoder and $G$-decoder, respectively, 
	$t_{<i}$ stands for the proceeding tokens before position $i$, and 
	$\Theta_{E}$, $\Theta_{U}$ and $\Theta_{G}$ denote the parameters of the shared-encoder, $U$-decoder and $G$-decoder, respectively.
	Since we deal with Chinese texts, we further adopt the \emph{whole word masking} strategy~\cite{cui2021pre} to sample word spans for masking (all the tokens from a sampled word will be masked), which is more suitable to capture the semantic information reflected by Chinese words. 
	
	\paratitle{Position-biased Masking}. 
	Unlike general texts, the mathematical text conveys important logic that threads the scattered evidences for deriving the final solution. 
	Intuitively, the contained tokens in a mathematical text are not of equal roles: the words at smaller positions are more likely to be an auxiliary evidence, and the words at larger positions are more likely to be an important hint to the answer\footnote{Since  our corpus contains the solution text, the tokens at larger positions of a mathematical text are nearer to the answer.}. Based on this idea, we employ a linearly increasing  weighting mechanism that assigns larger masking weights to words at larger positions as: 
	\begin{equation}
	s_i = \frac{30}{n-1} \times i,
	\label{eq-mask}
	\end{equation}
	where $ s_i$ is the sampling weight to be masked for the $i$-th position, $n$ is the sentence length, and $i$ is the position index ranging from 0 to $n-1$. With this weighting mechanism, the boundary words (\ie the first and last words) correspond to the masking probabilities of $0\%$ and  $30\%$, respectively, having an average probability of 15\% (equal to the original masking probability in BERT~\cite{devlin-etal-2019-bert}). In other words, the overall average masking probability  still remains as 15\%, while we re-distribute the sampling probabilities according to their positions. Such a masking strategy enforces the PLMs to focus more on the tokens that are nearer to the answer.

	\subsubsection{Advanced Course: Mathematical Logic Recovering}
	With the above token-level pre-training tasks,  PLM can grasp basic semantics about math symbols and description words. We further capture the mathematical  logic involving deductions and derivations  for  solving the problem, which is essential to the understanding of mathematical problems. In order to enhance the logic modeling ability of PLMs, we design two logic-related pre-training tasks by recovering shuffled sentences and formals, respectively. 
	
	\paratitle{Shuffled Sentences Recovering~(SSR)}. For the SSR task, we first shuffle the sentences in the solution text (denoted as $d$) for a given problem $q$, and reconstruct the shuffled text with the $G$-decoder. 
	Specifically, given a solution text $d$ consisting of multiple sentences, we randomly shuffle the order of these sentences to produce a corrupted solution text denoted as $\widetilde{d}_{S}$.
	The pre-training objective is to recover the original solution text $d$ based on $\widetilde{d}_{S}$, which can be formulated by minimizing the conditional language model loss by:
	\begin{equation}
	L_{SSR}=\sum_{i}-\log{p(t_i|t_{<i},  \widetilde{d}_{S};\Theta_{E}, \Theta_{G}} ),
	\label{eq-ssr}
	\end{equation}
	where we adopt an auto-regressive loss similar to BART~\cite{lewis-etal-2020-bart} to predict the original content. 
	
	\paratitle{Shuffled Formulas Recovering~(SFR)}. 
	In the mathematical text, formulas are the core elements that determine the underlying mathematical logic. Considering this point, we further extend the 
	above logic recovering task for the shuffled formulas.
	Specifically, we first shuffle the formulas $\{f_1, f_2, \cdots , f_m\}$ from the solution text to construct the corrupted solution text denoted as $\widetilde{d}_{F}$ and then recover the original content (only the formulas are shuffled).
	Such a pre-training task can be defined as:
	\begin{equation}
	L_{SFR}=\sum_{i}-\log{p(t_i|t_{<i}, \widetilde{d}_{F};\Theta_{E}, \Theta_{G}}).
	\label{eq-sfr}
	\end{equation}
	
	The two pre-training tasks are similar to the task of \emph{sentence order prediction} in BERT~\cite{devlin-etal-2019-bert}. However, we aim to capture the  logic underlying the mathematical text, instead of semantic relatedness between consecutive sentences.
	In our approach, we consider a more difficult task setting, where the PLM is required to recover the \emph{original content} (token-level prediction) instead of predicting the \emph{original order} (sentence-level prediction). 
	Such a pre-training task is essential to enhance the capacity of mathematical logic reasoning for PLMs. 
	Moreover, to avoid the catastrophic forgetting of basic courses, we  also incorporate the MLM and DAE tasks in Section~\ref{sec-MLM} as regularizers in this course.
	Finally, the logic-based pre-training objective is given by combining  the four losses as:
	\begin{equation}
	L_{AC}=L_{MLM}+L_{DAE}+L_{SSR}+L_{SFR}.
	\label{eq-ac}
	\end{equation}
	
	\subsubsection{Advanced Course: Solution Checking}
	The above token- and logic-based pre-training tasks can enhance the understanding capacity of  mathematical knowledge and logic to some extent. We further consider a  pre-training task that is directly related to the solving procedure. For mathematical problems, a minor mistake at some intermediate step will cause a failed solution.
	In order to derive the correct answer, the PLM should be able to 
	fully examine fine-grained evidences or elements in the 
	solving procedure. 
	Based on this idea, we design a pre-training task of dual-decoder correction, which mimics the problem solving process by human~\cite{huang2012learning}: a double check is usually required to verify the final solution. 
	In this pre-training course, we first utilize the two decoders  to generate the masked words or symbols, and then train them to detect and correct the generated errors from each other.
	
	\paratitle{Dual-Decoder Solution Checking~(SC)}. Specifically, given the solution text $d=\{t_1,t_2,\cdots,t_l\}$, we first mask a proportion of tokens  following the same strategy in Section~\ref{sec-MLM} to obtain the masked text $\widetilde{x}$.
	Then, we utilize the $U$-decoder and $G$-decoder to produce two recovered results based on their own generation probabilities of the masked tokens, denoted as $\widetilde{d}_{U}$ and $\widetilde{d}_G$.
	Next, we employ the two decoders to detect and correct the generated texts from each other, where the $\widetilde{d}_{G}$ and $\widetilde{d}_U$ will be examined and corrected by the $U$-decoder and $G$-decoder, respectively.
	In this way, the incorrect generations is expected to be corrected in a dual way. Such a pre-training process can be formulated as:
	\begin{align}
		L_{USC} &= \sum_{i}-\log{p(t_i|\widetilde{d}_G;\Theta_{E},\Theta_{U})}, \label{eq-usc}\\
		L_{GSC} &=\sum_{i}-\log{p(t_i|t_{<i},
			\widetilde{d}_U;\Theta_{E},\Theta_{G})}, 
		\label{eq-gsc}
	\end{align}
	where $L_{USC}$ and $L_{GSC}$ denote the self-correction losses for the $U$-decoder and $G$-decoder, respectively.
	These tasks can enhance the self-diagnosis capacity of the PLM by ensuring that all key points leading to the solution are correct, which is especially useful for complex mathematical problems requiring a long solving procedure.  
	Besides, we also incorporate the MLM and DAE tasks to prevent the forgetting of previous courses, and the learning objective of the solution checking course can be formulated as:
	\begin{equation}
	L_{SC}=L_{MLM}+L_{DAE}+L_{USC}+L_{GSC}.
	\label{eq-sc}
	\end{equation}

	To enhance the capacity of detecting incorrectly generated tokens, we enforce it to recover the entire sequence rather than the corrupted part only. 
	In this way, the model is required to distinguish whether a token is corrupted, and then only corrects the corrupted ones. To further reduce the influence of errors, the two decoders also perform self-checking on the generated sentences.
	Note that a problem might correspond to multiple solutions, here we only consider the sample answer in the solution text as the ground truth. In this setting, all the inconsistent generations are considered as incorrect predictions.  The rationale is that we only mask a small proportion of tokens so that the recovered solution should remain the same as original. We leave a more principled method to examine the correctness of the generated solutions in future work.   
	
	\subsection{Learning and Discussion}
	In this part, we present the learning and discussion of our model.
	
	\subsubsection{Learning}
	We adopt the CPT model~\cite{shao2021cpt} as the base architecture and utilize its pre-trained parameters to initialize the parameters of our model.
	We also randomly initialize the embeddings of math symbols that are out of the vocabulary of CPT.
	During pre-training, we design three courses to gradually learn the mathematical knowledge and logic from the mathematical corpus. 
	First, we train our model on the pre-training tasks that consist of MLM and DAE tasks using Eq.~\eqref{eq-mlm} and Eq.~\eqref{eq-dae} to learn basic mathematical semantics, where we incorporate the position-biased masking strategy to improve the adaptation to mathematical texts.
	Then, we adopt the logic-based course that recovers shuffled sentences and formulas using Eq.~\eqref{eq-ac} to improve the understanding of mathematical logic.
	Next, we employ the two decoders of our model to detect and correct the incorrect generations from each other using Eq.~\eqref{eq-sc}.
	Note that we also combine the MLM and DAE tasks in the latter two courses as regularizers to avoid the forgetting of basic mathematical semantics.
	Finally, for downstream tasks, we can fine-tune our model with task-specific datasets or learning objectives, and utilize the $U$-decoder or $G$-decoder for making the predictions.

	\subsubsection{Discussion}
	The focus of this paper is not to simply enhance the mathematical capacity by increasing the number of model parameters. Instead, we keep a base-sized PLM (about 121M parameters) for JiuZhang, and focus on designing more effective pre-training tasks that well adapt to Chinese mathematical texts. The major novelty of JiuZhang lies in the curriculum pre-training approach consisting of the basic course (masked token prediction) and the advanced courses (mathematical logic recovering and solution checking).
	Although existing methods~\cite{peng2021mathbert,gong-etal-2022-continual} attempt to adapt PLMs to math domain, they still rely on pre-training tasks as BERT to learn semantic representations for mathematical text. As a comparison, our proposed pre-training tasks are more capable of learning the essential knowledge and logic for mathematical  problem understanding, including position-biased masking, mathematical logic recovering and solution checking. Furthermore, we design a curriculum pre-training approach to scheduling the three kinds of tasks in a basic-to-advanced manner.
	Similar to the actual student learning process~\cite{laurillard1979processes},
	such a pre-training approach can help our model gradually understand and acquire mathematical knowledge and logic.
	
	\section{Experiments}
	To verify the effectiveness of our approach, we conduct experiments on nine tasks from high-school math education.
	
	\begin{table}[t]
		\centering
		\caption{Statistics of the datasets for nine evaluation tasks.}
		\begin{tabular}{clrrr}
			\bottomrule
			
			\textbf{Type} & \textbf{Task} & \textbf{Train} & \textbf{Dev} & \textbf{Test} \\ 
			\hline
			\multirow{3}*{Classification} & KPC & 8,721 & 991 & 1,985 \\
			& QRC & 10,000 & 2,000 & 4,000 \\
			& QAM & 14,000 & 2,000 & 4,000 \\
			\cline{1-5}
			\multirow{2}*{Retrieval} & SQR & 249,902 & 11,462 & 56,329 \\
			& QAR & 35,000 & 10,000 & 20,000 \\
			\hline
			\multirow{2}*{QA tasks} & MCQ & 
			22,000 & 3,982 & 7,466 \\
			& BFQ & 14,795 & 1,786 & 1,778 \\
			\cline{1-5}
			\multirow{2}*{Generation} & CAG & 14,795 & 1,786 & 1,778 \\
			& BAG & 16,000 & 1,976 & 1,977\\
			\bottomrule
		\end{tabular}
		\label{tab-static}
	\end{table}
	
	
	
	\subsection{Experimental Setup}
	
	\begin{table*}
		\centering
		\caption{Main results on three basic classification tasks and two math text retrieval tasks. The best and the second-best methods are denoted in bold and underlined fonts respectively. }
		\begin{tabular}{lcccccccccc}
			\bottomrule
			\textbf{Tasks} & \multicolumn{2}{c}{\textbf{KPC}} & \multicolumn{2}{c}{\textbf{QAM}} & \multicolumn{2}{c}{\textbf{QRC}} & \multicolumn{2}{c}{\textbf{SQR}} & \multicolumn{2}{c}{\textbf{QAR}}\\
			\cmidrule(r){1-1}\cmidrule(r){2-3}\cmidrule(r){4-5}\cmidrule(r){6-7}\cmidrule(r){8-9}\cmidrule(r){10-11}
			Metrics & Accuracy & F1-macro & Accuracy & F1-macro & Accuracy & F1-macro & HR@3 & NDCG@3 & HR@3 & NDCG@3 \\
			\hline
			TextCNN & 47.4 & 26.8 & 90.4 & 90.4 & 73.3 & 52.9 & 0.332 & 0.338 & 0.637 & 0.486 \\
			TextRCNN & 55.3 & 38.8 & 88.3 & 88.3 & 79.6 & 59.0 & 0.333 & 0.324 & 0.653 & 0.499 \\
			\hline
			BERT & 59.6 & 34.9 & 96.9 & 96.9 & 82.7 & 63.4 & 0.614 & 0.618 & 0.683 & 0.519 \\
			RoBERTa-wwm & 61.0 & 37.0 & 97.4 & 97.4 & 84.2 & 65.2 & 0.618 & 0.622 & 0.707 & 0.544 \\
			Mengzi & 56.6 & 29.5 & 97.3 & 97.3 & 81.7 & 62.8 & 0.625 & 0.625 & 0.672 & 0.520 \\
			BART & 62.7 & 41.9 & 97.7 & 97.6 & 82.0 & 63.0 & 0.610 & 0.610 & 0.688 & 0.530 \\
			CPT & 66.2 & 48.4 & 97.8 & 97.8 & 82.8 & 63.4 & 0.613 & 0.613 & 0.702 & 0.544 \\
			DAPT-BERT & 68.7 & 46.5 & 98.2 & 98.1 & 86.5 & 68.5 & 0.650 & 0.653 & 0.718 & 0.548 \\
			MathBert & 68.9 & 47.1 & 98.9 & 98.9 & 85.3 & 69.8 & 0.652 & 0.656 & 0.705 & 0.545 \\
			COMUS & 71.0 & \textbf{63.3} & 99.0 & 99.0 & 88.0 & 73.3 & 0.661 & 0.664 & \textbf{0.724} & \textbf{0.561}\\
			DAPT-CPT & \underline{72.0} & 58.0 & \underline{99.1} & \underline{99.1} & \underline{88.8} & \underline{76.7} & \underline{0.664} & \underline{0.668} & \underline{0.723} & \underline{0.556} \\
			\hline
			Ours & \textbf{73.3} & \underline{59.4} & \textbf{99.4} & \textbf{99.4} & \textbf{89.4} & \textbf{79.2} & \textbf{0.667} & \textbf{0.672} & \textbf{0.724} & \underline{0.556} \\
			\bottomrule
		\end{tabular}
		\label{tab-main-results}
	\end{table*}

	\begin{table*}
		\centering
		\caption{Main results on two question answering tasks and two analysis generation tasks. The best and the second-best methods are denoted in bold and underlined fonts respectively. }
		\begin{tabular}{lcccccccccc}
			\bottomrule
			\textbf{Tasks} & \textbf{MCQ} & \textbf{BFQ} & \multicolumn{4}{c}{\textbf{CAG}} & \multicolumn{4}{c}{\textbf{BAG}} \\
			\cmidrule(r){1-1}\cmidrule(r){2-2}\cmidrule(r){3-3}\cmidrule(r){4-7}\cmidrule(r){8-11}
			Metrics & Accuracy & Accuracy & BLEU-4 & ROUGE-2 & ROUGE-L & Accuracy & BLEU-4 & ROUGE-2 & ROUGE-L & Accuracy \\
			\hline
			Seq2Seq & 37.61 & 44.32 & 39.91 & 47.79 & 67.88 & 42.63 & \underline{39.86} & 48.15 & \underline{68.06} & 39.91 \\
			Transformer & 35.33 & 46.57 & \textbf{41.39} & 48.50 & 67.09 & 41.02 & \textbf{41.91} & 48.80 & 67.76 & \textbf{45.95}\\
			\hline 
			BART & 36.15 & 46.82 & 39.26 & 49.37 & 67.73 & 42.57 & 37.16 & 47.18 & 66.73 & 32.61 \\
			CPT & 37.90 & 46.31 & 39.56 & 50.07 & 68.20 & 43.34 & 37.20 & 47.99 & 67.25 & 34.26 \\
			DAPT-CPT & \underline{46.26} & \underline{53.41} & 40.46 & \underline{50.84} & \underline{68.87} & \underline{46.52} & 38.39 & \underline{49.28} & 68.04 & 41.82 \\
			\hline
			Ours & \textbf{47.73} & \textbf{54.60} & \underline{40.81} & \textbf{51.09} & \textbf{69.45} & \textbf{48.51} & 39.28 & \textbf{49.62} & \textbf{68.37} & \underline{44.03}\\
			\bottomrule
		\end{tabular}
		\label{tab-hard-results}
	\end{table*}

	\paratitle{Pre-training Corpus.} Our pre-training corpus is collected from Zhixue, a widely-used Chinese online education website, consisting of 1,276,952 high school math problems. Each question is associated with the problem type, problem statement and  solution text. 
	We first convert the formulas and math symbols into a unified LaTex format.
	Then, we preprocess these texts by simple regular expressions (removing extra spaces or unnecessary special symbols), and tokenize the Chinese sentences with the toolkit of Jieba\footnote{https://github.com/fxsjy/jieba} for whole word masking. More details can be found in Appendix~\ref{sec-pt-corpus}. 
	
	\paratitle{Evaluation Tasks.}
	We conduct nine evaluation tasks based on the collected math corpus, corresponding to different mathematical understanding and generation capacities, including three classification tasks, two retrieval tasks, two question answering tasks and two analysis generation tasks. 
	We split each task dataset into training/development/test sets. Note that the training set is used for fine-tuning. The statistics of these tasks are shown in Table~\ref{tab-static}.
	
	$\bullet$ \textbf{Knowledge Point Classification}~(KPC) is a multi-label classification task. Given a mathematical problem, 
	the task of KPC is to classify it into a set of pre-defined knowledge points. In this task, the knowledge points are created and annotated by professional teachers, and there are 387 knowledge points in total. 
	
	$\bullet$ \textbf{Question Relation Classification}~(QRC) is a six-class classification task. Given two math questions, the goal is to predict their relation from the given relation set: \{\emph{equivalent, similar, problem variant, conditional variant, situation variant, irrelevant}\}.
	
	$\bullet$ \textbf{Question-Answer Matching}~(QAM) is a binary classification task to predict if an answer is able to solve a given question. We sample an answer for each question as the negative example.
	
	$\bullet$ \textbf{Similar Question Retrieval}~(SQR) is a ranking task to retrieve similar questions. Given a question, it aims to rank the candidate questions based on their similarities with the query.
	
	$\bullet$ \textbf{Question-Answer Retrieval}~(QAR) is a ranking task to retrieve proper answers. Given a question, this task aims to rank  candidate answers based on their similarities with the question.
	
	$\bullet$ \textbf{Multiple-Choice Question Answering}~(MCQ) is to select the true answer from the multiple choices. The evaluation set is constructed with high-school multi-choice mathematical problems.
	
	$\bullet$ \textbf{Blank-Filling Question Answering}~(BFQ) is to generate the answer to fill the blank within the problem statement. We adopt high-school blank-filling mathematical problems for evaluation.
	
	$\bullet$ \textbf{Multiple-Choice Analysis Generation}~(CAG) aims to generate the answer and the proper analysis that explains the reasoning process given the question and its choices. 
	
	$\bullet$ \textbf{Blank-Filling Analysis Generation}~(BAG) aims to generate the answer and the proper analysis that explains the reasoning process for the blank-filling question.
	
	Since these tasks are in different difficulty levels, we divide them into two groups:  \underline{\emph{group $A$}} contains KPC, QAM, QRC, SQR and QAR, and  \underline{\emph{group $B$}} contains MCQ, BFQ, CAG and BAG.
	Overall, the tasks in group $B$ (question answering and generation tasks) are more difficult than those in group $A$ (classification and retrieval tasks).
	
	\paratitle{Evaluation Metrics.}
	For classification tasks (KPC, QAM and QRC), we adopt Accuracy and F1-macro as the evaluation metrics. For the retrieval tasks (SQR and QAR), we employ top-$k$ Hit Ratio (HR@$k$) and top-$k$ Normalized Discounted Cumulative Gain (NDCG@$k$). 
	Since the number of candidates is usually between 6 and 15, we adopt HR@3 and NDCG@3 for evaluation. 
	For question answering tasks (MCQ, BFQ), we adopt Accuracy for evaluation. For the BFQ task, we match the numerical value of the generated answer with the golden answer to avoid the format mismatch (\eg 1 and 1.0).
	For generation tasks (CAG, BAG), we use BLEU-4~\cite{Papineni2002BleuAM}, ROUGE-2 and ROUGE-L~\cite{Lin2004ROUGEAP} to evaluate the quality of the generated analysis. We also adopt Accuracy to evaluate the generated answers.
	
	\paratitle{Baseline Methods.}
	We compare our proposed approach with the following baseline methods, including non-pre-training methods, pre-training methods and continual pre-training methods:
	
	$\bullet$ \textbf{TextCNN}~\cite{kim-2014-convolutional} is a classic text representation model for text classification, using CNN on top of word vectors.
	
	$\bullet$ \textbf{TextRCNN}~\cite{lai2015recurrent} adopts RNN and CNN for classification tasks.
	
	$\bullet$ \textbf{Seq2Seq}~\cite{Bahdanau2015NeuralMT} consists of a GRU encoder and a GRU decoder with the attention mechanism for text generation tasks.
	
	$\bullet$ \textbf{Transformer}~\cite{vaswani2017attention} adopts the multi-head self-attention based encoder-decoder framework for generation tasks.
	
	$\bullet$ \textbf{BERT-Base}~\cite{devlin-etal-2019-bert} is pre-trained using the MLM task. We use the BERT-Base-Chinese checkpoint from  Huggingface\footnote{https://huggingface.co/bert-base-chinese}.
	
	$\bullet$ \textbf{BART-Base}~\cite{lewis-etal-2020-bart} is pre-trained by the denoising auto-encoding task. We utilize BART-Base-Chinese released by Shao~\etal~\cite{shao2021cpt}.
	
	$\bullet$ \textbf{RoBERTa-wwm}~\cite{cui2021pre} introduces the whole word masking strategy to mask whole words instead of individual Chinese characters.
	
	$\bullet$ \textbf{MathBERT}~\cite{peng2021mathbert} continually pre-trains BERT on a large-scale mathematical corpus, which revises the self-attention mechanism for encoding formulas and adopts MLM and formula-related tasks.
	
	$\bullet$ \textbf{CPT}~\cite{shao2021cpt} is a Chinese pre-trained language model consisting of a shared encoder, an understanding and a generation decoder. 
	
	$\bullet$ \textbf{Mengzi}~\cite{zhang2021mengzi} is a family of Chinese pre-trained language models. We adopt Mengzi-BERT for understanding tasks.
	
	$\bullet$ \textbf{DAPT-BERT}~\cite{gururangan-etal-2020-dont} conducts continual pre-training for BERT on domain-related corpus. We use the same pre-training mathematical corpus as our PLM, with the masked language model task. 
	
	$\bullet$ \textbf{DAPT-CPT} further pre-trains CPT on our collected pre-training corpus with the MLM and DAE tasks.
	
	$\bullet$ \textbf{COMUS}~\cite{gong-etal-2022-continual} adopts syntax-aware memory networks to capture structural semantics of math problems, and devises three pre-training tasks to further enhance the representations.
	
	\subsection{Main Results}
	The results of the comparison methods on the tasks of group $A$ and $B$ are shown in Table~\ref{tab-main-results} and Table~\ref{tab-hard-results}, respectively. 
	In general, non-pre-training methods perform worse than pre-training methods. The reason is that pre-training methods have captured more semantic knowledge from large-scale pre-training corpus.
	
	Among pre-training based baselines, CPT performs better than BERT, RoBERTa-wwm and BART mostly.
	A major reason is that CPT adopts an unbalanced architecture, consisting of a deep shared encoder and two shallow decoders for understanding and generation, respectively.
	Such a framework enables CPT to better learn general knowledge from multi-task pre-training and easily adapt to various tasks.
	As an exceptional case, BART outperforms CPT in CAG task, since its pre-training task is for text generation.
	
	Besides, we can see that continual pre-training methods (\ie MathBert, DAPT-BERT, COMUS and DAPT-CPT) outperform other baselines consistently. It is because mathematical texts contain important terminology and symbols, which can't be learned from the general corpus.
	As shown in \cite{gururangan-etal-2020-dont}, continual pre-training is an important technique that adapts PLM to specific domains or tasks, which is also useful to improve the understanding of mathematical problems.
	However, MathBERT does not perform well among the four methods. 
	It is likely because that its devised tasks focus on improving the understanding of a single math formula, while the test data from our corpus contains more formulas than the expected setting of MathBERT.
	In contrast, DAPT-CPT performs better than other baselines in most tasks. 
	As it adopts both the MLM and DAE tasks for continual pre-training, such a way helps DAPT-CPT better learn mathematical knowledge for math-related tasks.

	Finally, on most tasks, our model performs consistently better than baselines.
	With the specially designed curriculum pre-training strategy, our model can gradually capture mathematical knowledge and logic from basic and advanced pre-training courses.
	Such a curriculum pre-training strategy can improve the capacity of our model to solve various math-related tasks.
	However, Transformer outperforms our method in the BLEU-4 metric. After inspecting the results, we find  Transformer (without pre-training) generates more uninformative consecutive segments that match the reference text, but it can't capture important keywords or short phrases (lower ROUGE-2 and ROUGE-L).  
	
	\subsection{Ablation Study}
	\begin{figure*}[t!]
		\centering
		\includegraphics[width=\linewidth]{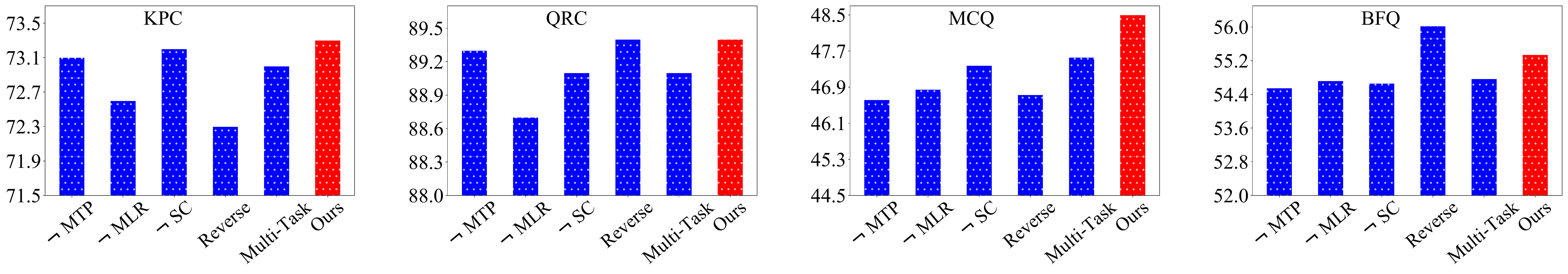}
		\caption{Ablation study of our approach on four tasks. 
			``$\neg$'' indicates that the corresponding course is removed in the curriculum pre-training stage, while the rest courses are kept. We abbreviate the courses of masked token prediction, mathematical logic recovering and solution checking as MTP, MLR and SC, respectively.}
		\label{fig-ablation}
	\end{figure*}
	
	As the major technical contribution, we design a curriculum pre-training approach consisting of three courses with different learning purposes.
	In this part, we examine the effect of each course on the performance of our model, by removing each course individually. Besides, as we schedule these courses in a basic-to-advanced order, it is important to examine how the pre-training order affects the final performance. For this purpose, we prepare two variants: (1) integrating the tasks within the above courses via a multi-task learning manner and (2) reversing the order of these courses.

	In Figure~\ref{fig-ablation}, we can see that removing any course or changing the curriculum order will mostly lead to a performance decrease on downstream tasks. It indicates the effectiveness of the current design for our curriculum pre-training approach. 
	Among these variants, the model performance significantly drops when we remove the MLR course (\ie mathematical logic recovering), which implies that it is important to consider mathematical logic in solving math tasks. 
	Besides, we can see that reversing the order of curriculum pre-training can outperform our approach in BFQ task but results in a performance drop in other tasks. 
	It indicates that the advanced-to-basic order might be useful to predict the masked tokens. 
	A possible reason is that it is more suited to the setting of BFQ task that predicts the blank of the given problems, but not other tasks.
	
	\subsection{Online $A/B$ Test}

	\begin{table}
		\centering
		\caption{Online $A/B$ test of our approach on the relevant problem recommendation task.}
		\begin{tabular}{lccc}
			\bottomrule
			& JiuZhang Wins & Baseline Wins \\
			\hline
			Ratio & 84.81 \% & 15.19\% \\
			\bottomrule
		\end{tabular}
		\label{table-online}
	\end{table}

	To further examine the effectiveness of our approach, we conduct the online $A/B$ test on Zhixue\footnote{\url{https://www.zhixue.com/}}, a personalized learning app that assists teachers to better teach students. It has over 45 million users in China. Specially, we conduct the test through the function of \emph{relevant problem recommendation~(RPR)} supported by this app. This function aims to recommend $k$ relevant problems ($k=3$ in this test) given a target problem. We obtain a small population of requests for comparing our approach (fine-tuned with the training data provided by this app) and the original  method deployed on the  app (a hybrid approach combining hand-crafted features and deep models). 
	Given a request, a user will be asked to select one better group of recommendation results based on her/his preference.

	We report the average winning ratios of the two methods in Table~\ref{table-online}.
	As we can see, our model performs better than the online deployed method. The major reason is that our model can produce better representations for mathematical problems (we use the simple inner product as the recommendation scores for our method), so that it can better model the similar relation between problems.

	\section{Conclusion}
	In this paper, we proposed JiuZhang, a Chinese pre-trained language model for mathematical problem understanding.
	We devised a set of pre-training tasks to effectively capture  mathematical knowledge and logic, and organized these tasks as the training curriculum to gradually learn the model parameters, from basic to advanced courses.
	In such a curriculum pre-training approach, we first learned basic semantics  about math symbols and established their semantic relatedness with text words,  then performed logic-based reasoning to recover shuffled sentences or formulas in the solution text, and finally conducted the difficult pre-training task of solution checking.
	Experimental results have shown that our approach outperforms several competitive baselines on nine math-related tasks, including basic classification tasks, text retrieval tasks, question answering tasks and analysis generation tasks.
	
	In future work, we consider incorporating more knowledge resources from the math domain (\eg the  explanation of knowledge points from math textbooks). Besides, we will also explore the scaling limit of our PLMs by increasing the parameter scale.

	\section*{Acknowledgement}
	This work was partially supported by Beijing Natural Science Foundation under Grant No. 4222027, and  National Natural Science Foundation of China under Grant No. 61872369, Beijing Outstanding Young Scientist Program under Grant No. BJJWZYJH012019100020098,
	and the Outstanding Innovative Talents Cultivation Funded Programs 2021. This work is also partially supported by Beijing Academy of Artificial Intelligence~(BAAI). Xin Zhao is the corresponding author.
	
	\bibliographystyle{ACM-Reference-Format}
	\bibliography{sample-base}
	
	\ignore{
		\begin{table*}[t!]
			\centering
			\small
			\caption{An examples of the pre-training data after processing.}
			\begin{tabular}{|l|l|} 
				\hline
				\textbf{Problem Statement} & 设 0 \textless \textbackslash{}\textbackslash{}alpha \textless \textbackslash{}\textbackslash{}pi, 且 \textbackslash{}\textbackslash{}sin \textbackslash{}\textbackslash{}frac \{ \textbackslash{}\textbackslash{}alpha \} \{ 2 \} = \textbackslash{}\textbackslash{}frac \{ \textbackslash{}\textbackslash{}sqrt \{ 3 \} \} \{ 3 \} . 则 \textbackslash{}\textbackslash{}sin \textbackslash{}\textbackslash{}alpha = \textbackslash{}\textbackslash{}LongUnderLine . \\ 
				\hline
				\multirow{5}{*}{\textbf{Solution Text}} & 解~~0 \textless \textbackslash{}\textbackslash{}alpha \textless \textbackslash{}\textbackslash{}pi, 且 \textbackslash{}\textbackslash{}sin \textbackslash{}\textbackslash{}frac \{ \textbackslash{}\textbackslash{}alpha \} \{ 2 \} = \textbackslash{}\textbackslash{}frac \{ \textbackslash{}\textbackslash{}sqrt \{ 3 \} \} \{ 3 \} , 可得 \textbackslash{}\textbackslash{}cos \textbackslash{}\textbackslash{}frac \{ \textbackslash{}\textbackslash{}alpha \} \{ 2 \} = \textbackslash{}\textbackslash{}sqrt \{ 1 - \\
				& 
				( \textbackslash{}\textbackslash{}frac \{ \textbackslash{}\textbackslash{}sqrt \{ 3 \} \} \{ 3 \} ) \^{} \{ 2 \} \} = \textbackslash{}\textbackslash{}frac \{ \textbackslash{}\textbackslash{}sqrt \{ 6 \} \} \{ 3 \} ~. \textbackslash{}\textbackslash{}sin \textbackslash{}\textbackslash{}alpha = 2 \textbackslash{}\textbackslash{}sin \textbackslash{}\textbackslash{}frac \{ \textbackslash{}\textbackslash{}alpha\} \{ 2 \}\textbackslash{}\textbackslash{}cos~\textbackslash{}\textbackslash{}frac ~ \\
				&
				\{ \textbackslash{}\textbackslash{}alpha\} \{ 2 \} = 2 \textbackslash{}\textbackslash{}times~\textbackslash{}\textbackslash{}frac \{ \textbackslash{}\textbackslash{}sqrt \{ 3 \} \} \{ 3 \} \textbackslash{}\textbackslash{} times~\textbackslash{}\textbackslash{}frac \{ \textbackslash{}\textbackslash{}sqrt \{ 6 \} \} \{ 3 \} = \textbackslash{}\textbackslash{}frac \{ 2 \textbackslash{}\textbackslash{}sqrt \{ 2 \} \}
				\{ 3 \}. 故答案为~ \\
				&
				\textbackslash{}\textbackslash{}frac \{ 2 \textbackslash{}\textbackslash{}sqrt \{ 2 \} \} \{ 3 \} . 利用同角三角函数基本关系式求解余弦函数 , 然后利用二倍角公式求解即可 . 本题考查\\
				&
				二倍角公式的应用 , 同角三角函数基本关系式的应用 , 考察计算能力 . \\
				\hline
			\end{tabular}
			\label{table-pretraining-case}
		\end{table*}
	}
	
	\newpage
	\appendix
	
	
	\section{Pre-training Data Collection}
	
	\label{sec-pt-corpus}
	The dataset supplied by iFLYTEK Co., Ltd. was collected from the Chinese education website Zhixue.
	The raw data is organized in the format of HTML webpage. To extract the useful information, we parsed the HTML webpage with regular regressions, to obtain the problem statements and solution text.
	However, the contained math symbols and formulas within crawled mathematical problems aren't  in a unified format (\eg $x**y$ and $x^y$).
	We invited professionals and designed standards to unify the math symbols and operators within the above data.
	Next, we filtered low-quality examples with unidentifiable information or overlength text. 
	We also removed the overlapping data from the corpus with the test sets of downstream tasks for a fair comparison.
	Finally, we obtained 1,276,952 examples for pre-training (see a sample preprocessed mathematical text in Table~\ref{table-pretraining-case}).
	
	\begin{CJK*}{UTF8}{gbsn}
		\begin{table}[h]
			\centering
			\small
			\caption{A sample  mathematical text after preprocessing.}
			\begin{tabular}{|c|} 
				\hline
				\textbf{Problem Statement}\\ 
				\hline
				\multicolumn{1}{|l|}{\begin{tabular}[c]{@{}l@{}}设 0 \textless \textbackslash{}\textbackslash{}alpha \textless \textbackslash{}\textbackslash{}pi, 且 \textbackslash{}\textbackslash{}sin \textbackslash{}\textbackslash{}frac \{ \textbackslash{}\textbackslash{}alpha \} \{ 2 \} = \textbackslash{}\textbackslash{}frac \{~\textbackslash{}\textbackslash{}sqrt~\{~3 \}~\}\\\{ 3 \} .~则 \textbackslash{}\textbackslash{}sin~\textbackslash{}\textbackslash{}alpha = \textbackslash{}\textbackslash{}LongUnderLine .\end{tabular}}\\ 
				\hline
				\textbf{Solution Text}\\ 
				\hline
				\multicolumn{1}{|l|}{\begin{tabular}[c]{@{}l@{}}解 0 \textless \textbackslash{}\textbackslash{}alpha \textless \textbackslash{}\textbackslash{}pi, 且 \textbackslash{}\textbackslash{}sin \textbackslash{}\textbackslash{}frac \{ \textbackslash{}\textbackslash{}alpha \} \{ 2 \} = \textbackslash{}\textbackslash{}frac ~\{\textbackslash{}\textbackslash{}sqrt~\{~3 \}~\}\\\{ 3 \} ,~可得~\textbackslash{}\textbackslash{}cos \textbackslash{}\textbackslash{}frac \{ \textbackslash{}\textbackslash{}alpha \} \{ 2 \} =~\textbackslash{}\textbackslash{}sqrt \{ 1 - (\textbackslash{}\textbackslash{}frac~\{ \textbackslash{}\textbackslash{}sqrt~\{ 3 \} \}\\ \{ 3 \} )~\^{}~\{2 \} \} = \textbackslash{}\textbackslash{}frac\{~\textbackslash{}\textbackslash{}sqrt \{ 6 \} \} \{ 3 \} .~\textbackslash{}\textbackslash{}sin~\textbackslash{}\textbackslash{}alpha = 2 \textbackslash{}\textbackslash{}sin~\textbackslash{}\textbackslash{}frac \{ \\ \textbackslash{}\textbackslash{}alpha \}~\{ 2 ~\} \textbackslash{}\textbackslash{}cos~\textbackslash{}\textbackslash{}frac \{~\textbackslash{}\textbackslash{}alpha \}~\{ 2~\} = 2 \textbackslash{}\textbackslash{}times~\textbackslash{}\textbackslash{}frac \{\textbackslash{}\textbackslash{}sqrt~\{ 3 \} \}\\\{ 3 \} \textbackslash{}\textbackslash{} times~\textbackslash{}\textbackslash{}frac \{ \textbackslash{}\textbackslash{}sqrt \{ 6 \}~\}~\{ 3 \} = \textbackslash{}\textbackslash{}frac~\{ 2~\textbackslash{}\textbackslash{}sqrt~\{ 2 \} ~\}\{ 3 \} 故答\\案为~\textbackslash{}\textbackslash{}frac \{ 2 \textbackslash{}\textbackslash{}sqrt \{ 2~\} \} \{ 3 \} .\end{tabular}}  \\
				\hline
			\end{tabular}
			\label{table-pretraining-case}
		\end{table}
	\end{CJK*}
	
	\begin{table}[t!]
		\small
		\centering
		\caption{Parameter settings of the baselines.}
		\label{tab-baseline}
		\begin{tabular}{|c|c|} 
			\hline
			\textbf{Models} & \textbf{Settings}                    \\ 
			\hline
			\multicolumn{2}{|c|}{Classification and Retrieval Task} \\ 
			\hline
			TextCNN~\&~TextRCNN & \begin{tabular}[c]{@{}c@{}}AdamW, learning\_rate=0.0001\\warmup\_ratio=0.05\\batch\_size=32\\\end{tabular} \\
			\hline
			\begin{tabular}[c]{@{}c@{}}BERT~\&~RoBERTa-wwm~\&~Mengzi\\\&~BART~\&~CPT~\&~MathBERT\\\&~DAPT-BERT~\&~DAPT-CPT\end{tabular} & \begin{tabular}[c]{@{}c@{}}AdamW, learning\_rate=0.00003\\warmup\_ratio=0.05\\batch\_size=32\end{tabular} \\
			\hline
			\multicolumn{2}{|c|}{QA and Generation Task} \\ 
			\hline
			Seq2Seq~\&~Transformer & \begin{tabular}[c]{@{}c@{}}AdamW, learning\_rate=0.001\\warmup\_ratio=0.05\\batch\_size=64\\\end{tabular} \\ 
			\hline
			BART~\&~CPT~\&~DAPT-CPT & \begin{tabular}[c]{@{}c@{}}AdamW, learning\_rate=0.00005\\warmup\_ratio=0.1\\batch\_size=64\end{tabular} \\
			\hline
		\end{tabular}
	\end{table}

	\section{Implementation Details.}
	In our approach, the numbers of Transformer layers in the shared encoder, $U$-decoder and $G$-decoder are set to 10, 2, and 2, respectively.
	For each course, we pre-train the parameters with a batch size of 256 for 100,000 steps. The max length of input sequences is set as 512. 
	We use AdamW \cite{loshchilov2018decoupled} optimization with the learning rate of 3e$^{-5}$, and warmup the learning rate for the first 5,000 steps then decay the weight with a ratio of 0.01.
	During fine-tuning, we use AdamW with the same setting as pre-training.
	For classification and retrieval tasks, we utilize the shared encoder and the $U$-decoder for classification. We set the initial learning rate as 3e$^{-5}$ and batch size as 32.
	For QA and generation tasks, we utilize the shared encoder with the $G$-decoder to accomplish them in a seq2seq fashion. Texts are tokenized with a maximum length of 512 and 128 for input and output, respectively. We set the initial learning rate as 5e$^{-5}$ and batch size as 64.
	We perform the curriculum pre-training on 8 RTX 3090 24G GPUs.
	During fine-tuning, we construct the model inputs of all downstream tasks as:
	
	\textbf{QRC and SQR:} [CLS] $q_1$ [SEP] $q_2$ [SEP].
	
	\textbf{QAM:}  [CLS] $q$ [SEP] $a$ [SEP].
	
	\textbf{QAR:} [CLS] $q$ [SEP] and [CLS] $a$ [SEP].
	
	\textbf{KPC, MCQ, BFQ, CAG and BAG:} [CLS] $q$ [SEP]. 
	
	For baselines, we implement them based on HuggingFace Transformers\footnote{https://huggingface.co/transformers/}.
	We report the parameter settings of baseline models used throughout the experiments in Table~\ref{tab-baseline}, and present the procedure of our curriculum pre-training in Algorithm~\ref{al_pt}.

	\begin{algorithm}
		\caption{The curriculum pre-training algorithm.}\label{al_pt}
		\small
		\SetKwData{Left}{left}\SetKwData{This}{this}\SetKwData{Up}{up}
		\SetKwFunction{Union}{Union}\SetKwFunction{Sample}{Sample}\SetKwFunction{MaskMLM}{MaskMLM}\SetKwFunction{MaskDAE}{MaskDAE}
		\SetKwInOut{Input}{Input}\SetKwInOut{Parameter}{Parameter}

		\Input{mathematical text corpus $X=\{(q,d)\}$}
		\Parameter{The parameters of the encoder $\Theta_{E}$, $U$-decoder $\Theta_{U}$, $G$-decoder $\Theta_{G}$, steps of the three courses $M_1$, $M_2$, $M_3$}
		\BlankLine
		\tcp{Basic Course: Masked Token Prediction}
		\For{$t\leftarrow 1$ \KwTo $M_1$}{
			Randomly sample half of examples from the batch as $X^U_{t}$ and the other half as $X^G_{t}$, and perform masking\;
			Pre-train $\Theta_{E}$ and $\Theta_{U}$ on $X^U_{t}$ by MLM task using Eq.~\eqref{eq-mlm}\;
			Pre-train $\Theta_{E}$ and $\Theta_{G}$ on $X^G_{t}$ by DAE task using Eq.~\eqref{eq-dae}\;
		}
		\tcp{Advanced Course: Mathematical Logic Recovering}
		\For{$t\leftarrow 1$ \KwTo $M_2$}{
			Pre-train $\Theta_{E}$, $\Theta_{U}$ and $\Theta_{G}$ by MLM and DAE tasks as lines 2-5\;
			Sample half of examples from the batch as $X^S_{t}$ and the others as $X^F_{t}$, and shuffle the sentences and formulas, respectively\;
			Pre-train $\Theta_{E}$ and $\Theta_{G}$ on $X^S_{t}$ by SSR task using Eq.~\eqref{eq-ssr}\;
			Pre-train $\Theta_{E}$ and $\Theta_{G}$ on $X^F_{t}$ by SFR task using Eq.~\eqref{eq-sfr}\;
		}
		\tcp{Advanced Course: Solution Checking}
		\For{$step\leftarrow 1$ \KwTo $M_3$}{
			Pre-train $\Theta_{E}$, $\Theta_{U}$ and $\Theta_{G}$ by MLM and DAE tasks as lines 2-5, and obtain the generated text from $U$-decoder and $G$-decoder as $\widetilde{d_U}$ and $\widetilde{d_G}$, respectively\;
			Pre-train $\Theta_{E}$ and $\Theta_{U}$ on $\widetilde{d_U}$ by USC loss using Eq.~\eqref{eq-usc}\;
			Pre-train $\Theta_{E}$ and $\Theta_{G}$ on $\widetilde{d_G}$ by GSC loss using Eq.~\eqref{eq-gsc}\;
		}
	\end{algorithm}\DecMargin{1em}
	
	\section{Tuning the Number of Pre-Training Steps}
	
	\begin{CJK*}{UTF8}{gbsn}
		\begin{table*}
			\caption{Case study on the MCQ task.}
			\begin{tabular}{|l|ll|}
				\hline
				\textbf{Math Problem} &
				\multicolumn{2}{l|}{\begin{tabular}[c]{@{}l@{}}已知 $sin ( \alpha + \frac { \pi }{ 6 }) = \frac { 2 \sqrt { 5 }}{ 5 } $,则 $\cos ( \frac { \pi }{ 3 } - \alpha ) = ( )$ \\ 
						It is known that $sin ( \alpha + \frac { \pi }{ 6 }) = \frac { 2 \sqrt { 5 }}{ 5 } $,then $\cos ( \frac { \pi }{ 3 } - \alpha ) = ( )$ \\
						A. $\frac { \sqrt { 5 } } { 5 }$  B.$- \frac { \sqrt { 5 } } {5 } $ C. $\frac { 2 \sqrt {5 } } {5}   $   D.$- \frac { 2 \sqrt { 5 } }{ 5 }$ \\
				\end{tabular}} \\ \hline
				\textbf{Ground-Truth Analysis} &
				\multicolumn{2}{l|}{\begin{tabular}[c]{@{}l@{}}
						$\because$ $\cos ( \frac { \pi } { 3 } - \alpha ) = \cos {[} \frac { \pi } { 2 } – ( \alpha + \frac { \pi } { 6 } ) {]} = \sin ( \alpha + \frac { \pi } { 6 } ) = \frac { 2 \sqrt { 5 } } { 5 }$ . 故选C \\
						$\because$ $\cos ( \frac { \pi } { 3 } - \alpha ) = \cos {[} \frac { \pi } { 2 } – ( \alpha + \frac { \pi } { 6 } ) {]} = \sin ( \alpha + \frac { \pi } { 6 } ) = \frac { 2 \sqrt { 5 } } { 5 }$. Therefore, choose C
				\end{tabular}} \\ \hline
				\textbf{DAPT-CPT} &
				\multicolumn{2}{l|}{\begin{tabular}[c]{@{}l@{}}
						$ \cos( \frac { \pi } { 3 } - \alpha ) = \sin \alpha = - \sqrt { 5 } $ , 又因为 $\alpha$为锐角，所以 $ \alpha = \frac { b } { a } $，故选C \\
						$ \cos( \frac { \pi } { 3 } - \alpha ) = \sin \alpha = - \sqrt { 5 } $, and because $\alpha$ is an acute angle,  we have $ \alpha = \frac { b } { a } $, so choose C
				\end{tabular}} \\ \hline
				\textbf{Ours} &
				\multicolumn{2}{l|}{\begin{tabular}[c]{@{}l@{}}
						$ \cos ( \frac { \pi } { 3 } - \alpha ) = \{ \cos [ \pi + \frac { \pi } { 3 } - \alpha ] = \sin ( \alpha + \frac{ n } { 6 } ) = \frac { 2 \sqrt { 5 } } {5} $ 故选C \\
						$ \cos ( \frac { \pi } { 3 } - \alpha ) = \{ \cos [ \pi + \frac { \pi } { 3 } - \alpha ] = \sin ( \alpha + \frac{ n } { 6 } ) = \frac { 2 \sqrt { 5 } } {5} $ . Therefore, choose C
				\end{tabular}} \\ \hline \hline
				\textbf{Math Problem} &
				\multicolumn{2}{l|}{\begin{tabular}[c]{@{}l@{}}
						若复数 $ z = \frac { 2 i } { 1 - i } $ ( \{ i \} 是虚数单位 ), 则z的共轭复数 $ \overline { z } = ( ) $  \\ 
						If the complex number $ z = \frac { 2 i } { 1 - i } $ ( \{ i \} is the imaginary unit ),  then the complex conjugate $ \overline { z } = ( ) $ \\
						A. $ 1 + { i } $   B. $ 1 - { i } $  C. $ - 1 + { i } $   D. $ - 1 - { i } $ \end{tabular}} \\ \hline
				\textbf{Ground-Truth Analysis} &
				\multicolumn{2}{l|}{\begin{tabular}[c]{@{}l@{}}
						复数$ z = \frac { 2 i } { 1 - i } = \frac { 2 i ( 1 + i ) } { ( 1 - i ) \cdot ( 1 + i ) } = - 1 + i $, 则z的共轭复数 $ \overline { z } = - 1 - i $. 故选D \\
						Complex number $ z = \frac { 2 i } { 1 - i } = \frac { 2 i ( 1 + i ) } { ( 1 - i ) \cdot ( 1 + i ) } = - 1 + i $ ,  then the complex conjugate $ \overline { z } = - 1 - i $. Therefore, choose D
				\end{tabular}} \\ \hline
				\textbf{DAPT-CPT} &
				\multicolumn{2}{l|}{\begin{tabular}[c]{@{}l@{}}
						$ z = \frac { 2 i ( 1 + i ) } { 2 } = i + 1 $, 则z 的共轭复数 $ \overline { z } = - 1 - i $. 故选D \\
						$ z = \frac { 2 i ( 1 + i ) } { 2 } = i + 1 $, then the complex conjugate $ \overline { z } = - 1 - i $. Therefore, choose D
				\end{tabular}} \\ \hline
				\textbf{Ours} &
				\multicolumn{2}{l|}{\begin{tabular}[c]{@{}l@{}}
						复数 $ z = \frac { 2 i } { 1 - i } = \frac { ( 2 i ) ( 1 + i ) } { ( 1 - i ) ( 1 + i ) } =  1 + i $, 则复数的共轭复数$ \overline { z } = - 1 - \boldsymbol { i } $. 故选D \\
						Complex number $ z = \frac { 2 i } { 1 - i } = \frac { ( 2 i ) ( 1 + i ) } { ( 1 - i ) ( 1 + i ) } =  1 + i $ then the complex conjugate $ \overline { z } = - 1 - \boldsymbol { i } $. Therefore, choose D
				\end{tabular}} \\ \hline
			\end{tabular}
			\label{case-study}
		\end{table*}
	\end{CJK*}
	
\begin{figure}
		\centering
		\includegraphics[width=0.45\textwidth]{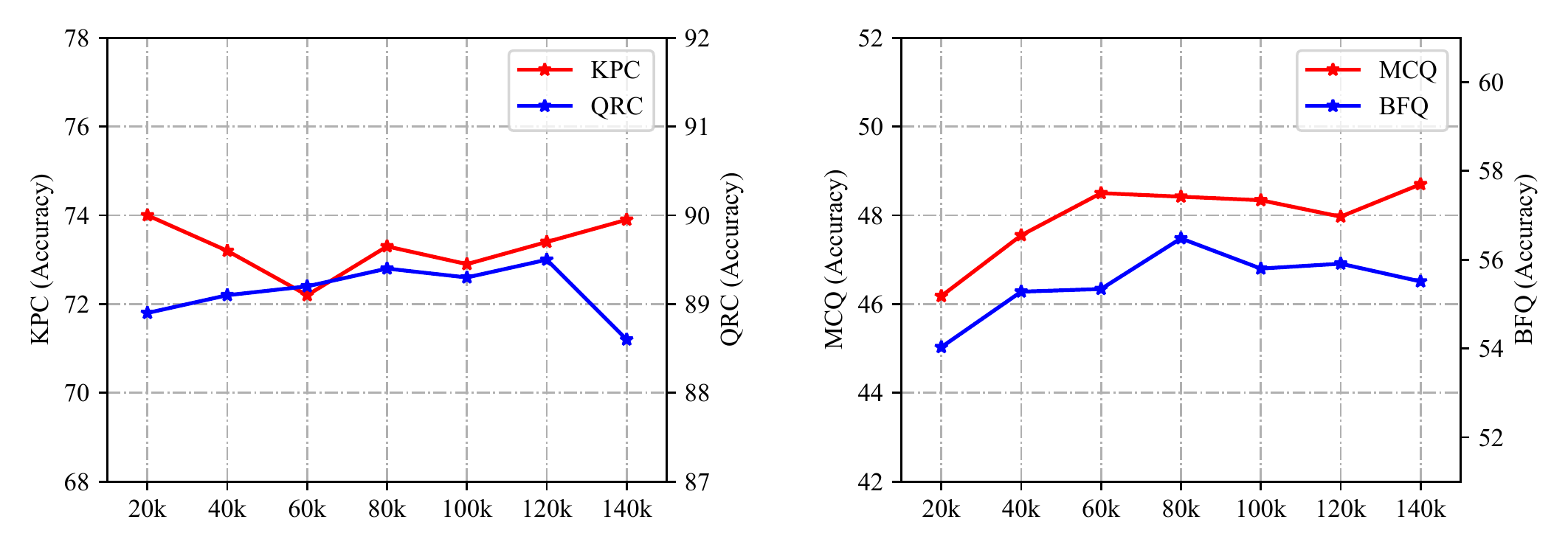}
		\caption{Performance tuning \emph{w.r.t.} the pre-training steps ($T$). }
		\label{fig-steps}
		\vspace{-0.2cm}
	\end{figure}

\begin{figure}
		\centering
		\includegraphics[width=0.45\textwidth]{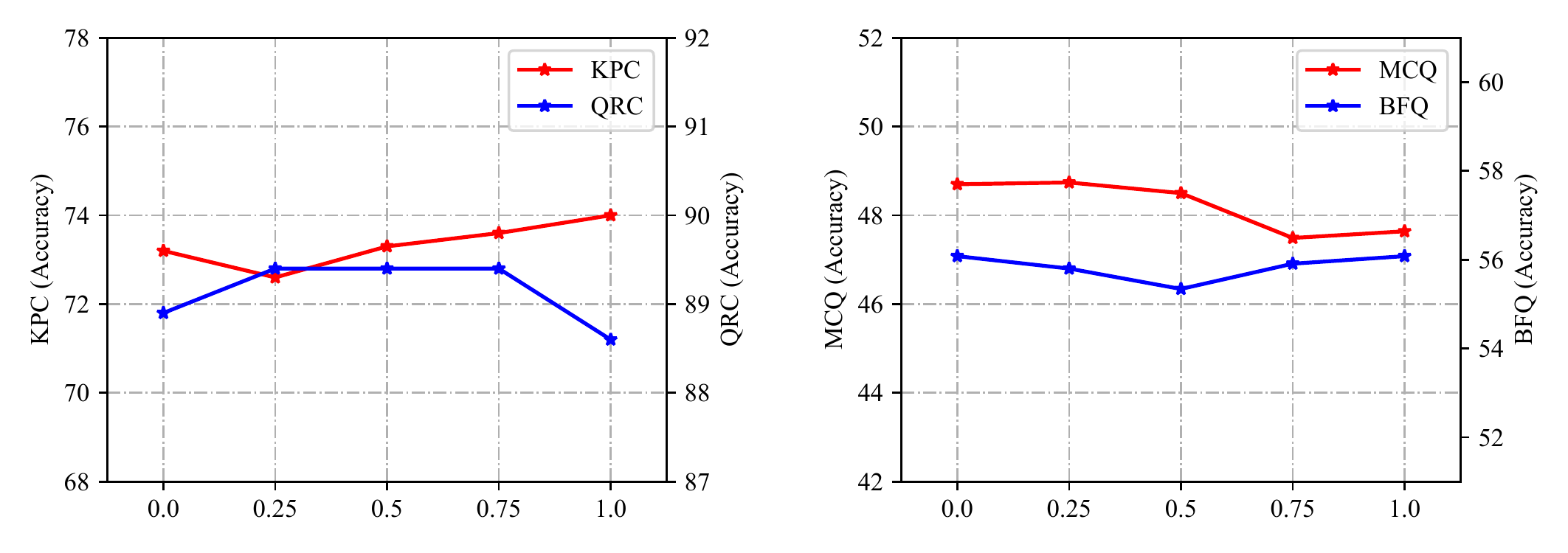}
		\caption{Performance tuning \emph{w.r.t.} the training step ratio ($\gamma$).}
		\label{fig-rate}
		\vspace{-0.2cm}
	\end{figure}

	For pre-trained language models,
	it is important to conduct a sufficient number of pre-training steps to 
	achieve the desired performance.
	Here, we examine how the model performance changes \emph{w.r.t.} the number of pre-training steps, denoted by $T$, which can provide important practical guidelines for reproducible implementation. Besides, since we have two different decoders, we need to schedule the number of pre-training steps assigned to each decoder. For this purpose, we further incorporate a pre-training step ratio (denoted as $\gamma$) to distribute the training steps  to the two decoders: $\gamma \cdot T$ for $U$-decoder and  $(1-\gamma) \cdot T$ for $G$-decoder. 
	We conduct the performance tuning experiments on the KPC, QRC, MCQ and BFQ tasks and show the performance curves of Accuracy in Figure~\ref{fig-steps} (tuning $T$) and  Figure~\ref{fig-rate} (tuning $\gamma$).

	First, in Figure~\ref{fig-steps}, we can observe that our model achieves the best performance with $140K$ training steps for KPC and MCQ tasks while with $80K$ training steps for BFQ tasks.
	It indicates that different tasks may require different numbers of pre-training steps to achieve the optimal performance: more training steps are required for KPC and MCQ tasks than the BFQ task.
	\ignore{\textcolor{blue}{it might be because too many pre-training steps may lead to overfitting on the MLM and DAE objectives that has a negative influence on the BFQ tasks with a similar blank-filling setting.}}
	
	Besides,  the ratio $\gamma$ controls the number of training steps scheduled for each decoder. As we can see from  Figure~\ref{fig-rate}, the best performance for  the tasks of KPC, MCQ and BFQ occur with extreme cases of  $\gamma$ (either 0 or 1). The major reason is that the two decoders are designed for understanding and generation, respectively, and it is intuitive that we should train $G$-decoder ($U$-decoder)  with more steps  for the generation (understanding) task. In practice, we have to deal with both generation and understanding tasks, by making a trade-off between the two tasks via $\gamma$.

	\section{Case Study}
	
	In this part, we display two sample analyses generated by our model from the MCQ task for a case study.
	We also show the analyses generated by the best baseline DAPT-CPT as a comparison.
	These cases are shown in Table~\ref{case-study}.
	
	First, we can see that DAPT-CPT has made mistakes such as calculation error and improper deduction.
	These minor mistakes are intolerant in solving mathematical problems and may lead to wrong predictions of answers.
	As a comparison, our model can generate a more complete analysis with clear logic for selecting the correct answer from the given choices. 
	Such an inference process is similar to the ground-truth one produced by experts.
	Besides, it is interesting to see that our model can successfully accomplish the deduction of the formals and obtain the true answer.
	It indicates that our model can effectively deal with the math symbols with proper mathematical logic.
	These results show that it is promising to design proper AI models to solve mathematical problems and also benefit the education of high-school students.

\end{document}